\documentclass[10pt,twocolumn,letterpaper]{article}

\usepackage{cvpr}

\definecolor{iccvblue}{rgb}{0.21,0.49,0.74}
\usepackage[pagebackref,breaklinks,colorlinks,allcolors=iccvblue]{hyperref}

\usepackage{amsmath,bm}
\usepackage{multirow}

\usepackage{graphicx}
\usepackage{booktabs}
\usepackage{pgfplots}

\def\paperID{***} 
\def\confName{CVPRW}
\def\confYear{2026}

\title{RetroMotion: Retrocausal Motion Forecasting Models are Instructable}

\author{
Royden Wagner$^{1,2}$\hspace{3mm}
{\"O}mer~{\c{S}}ahin~Ta{\c{s}}$^{1,2}$\hspace{3mm}
Felix Hauser$^{2}$\hspace{3mm}
Marlon Steiner$^{1}$\hspace{3mm}
Dominik Strutz$^{1}$
\and
Abhishek Vivekanandan$^{2}$\hspace{3mm}
Jaime Villa$^{3}$\hspace{3mm}
Yinzhe Shen$^{1}$\hspace{3mm}
Carlos Fernandez$^{1}$\hspace{3mm}
Christoph Stiller$^{1,2}$\vspace{2mm}\\ 
  $^{1}$Karlsruhe Institute of Technology \hspace{1mm} $^{2}$FZI Research Center for Information Technology \\
  $^{3}$University Charles III of Madrid
}

\begin{document}
\maketitle

\begin{abstract}
Motion forecasts of road users (i.e., agents) vary in complexity depending on the number of agents, scene constraints, and interactions.
In particular, the output space of joint trajectory distributions grows exponentially with the number of agents.
Therefore, we decompose multi-agent motion forecasts into (1) marginal  distributions for all modeled agents and (2) joint distributions for interacting agents.
Using a transformer model, we generate joint distributions by re-encoding marginal distributions followed by pairwise modeling.
This incorporates a retrocausal flow of information from later points in marginal trajectories to earlier points in joint trajectories.
For each time step, we model the positional uncertainty using compressed exponential power distributions.
Notably, our method achieves strong results in the Waymo Interaction Prediction Challenge and generalizes well to the Argoverse 2 and V2X-Seq datasets.
Additionally, our method provides an interface for issuing instructions.
We show that standard motion forecasting training implicitly enables the model to follow instructions and adapt them to the scene context. \vspace{2mm}\newline
\textbf{GitHub repository:} \url{https://github.com/kit-mrt/future-motion}

\end{abstract}

\section{Introduction}
Motion in traffic scenarios ranges from complex, interactive behaviors among road users in urban environments to more uniform trajectories on highways.
Data-driven methods for motion forecasting \citep{ngiam2022scene, shi2022motion, cui2023gorela} address this with multiple choice learning \citep{guzman2012multiple, lee2016stochastic}, where choices are trajectories of future positions.
In complex scenarios with many road users (i.e., agents) and choices, the distribution over future trajectories is likely multimodal. 
A common simplification is to model marginal distributions per agent \citep{nayakanti2023wayformer, zhou2023query, zhang2024real}.
To improve interaction modeling, recent methods predict joint distributions over multiple agents \citep{ngiam2022scene, seff2023motionlm, jiang2023motiondiffuser}.
However, when modeling the joint distribution over all agents in a scenario, the output space grows exponentially with the number of agents.
Therefore, we decompose motion forecasts into marginal forecasts for non-interacting agents and joint forecasts for interacting agents.
Specifically, we train a shared scene encoder and two decoders for marginal and joint motion forecasting.
We generate joint trajectory distributions by re-encoding marginal distributions followed by pairwise modeling.
Our two-stage decoding mechanism incorporates a \textit{retrocausal} flow of information from later points in marginal trajectories to earlier points in joint trajectories.
This lowers the modeling burden on initial marginal forecasts and provides an interface for issuing instructions through trajectory modifications.
We show that regular training of motion forecasting leads to the ability to follow goal-based instructions and to adapt basic directional instructions to the scene context.

Recent planning algorithms for self-driving vehicles \citep{tacs2023decision, geisslinger2023ethical, bouzidi2024motion} benefit from probabilistic uncertainty estimates.
Therefore, we train our decoders to model positional uncertainty using maximum likelihood estimation.
While related methods assume either normal \citep{shi2022motion, shi2024mtr++, zhang2024real} or Laplace distributions \citep{zhou2023query, zhou2023qcnext}, our decoders predict variably shaped exponential power distributions.
Our experiments show that learned distribution shapes tend to be Laplace-like, yet outperform plain Laplace distributions in terms of forecasting accuracy.
In addition, we reduce computational complexity by compressing the location parameters of distributions using discrete cosine transforms.


Our main contributions are:
\begin{enumerate}
    \item Our method connects marginal and joint motion forecasts through a retrocausal flow of information. This reduces the modeling burden on initial marginal forecasts and enables us to issue instructions by modifying marginal forecasts (see \Cref{fig:directional}). 
    \item We model positional uncertainty with compressed exponential power distributions, resulting in higher forecasting accuracy than with normal or Laplace distributions (see \Cref{subsection:analyzing_reps}).
\end{enumerate}
\section{Related work}


\textbf{Marginal motion forecasting}
methods model future motion per agent using marginal distributions of trajectories \citep{nayakanti2023wayformer, zhou2023query, zhang2024real}.
Related methods extend marginal forecasting models with auxiliary goal prediction \cite{gilles2022thomas} and dense prediction objectives \cite{shi2022motion, shi2024mtr++} or perform conditional \cite{sun2022m2i, rowe2023fjmp} and joint forecasting \cite{luo2023jfp} in a post hoc manner.
With auxiliary goal prediction, interaction modeling is more implicit than with joint modeling.
In post hoc methods, the flow of information from marginal trajectories to joint trajectorise is limited since they are first decoded and aggregated.
Specifically, the trajectory aggregation is non-differentiable non-maximum suppression (see \citep{luo2023jfp}).

\textbf{Joint motion forecasting}
methods model future motion with joint distributions of trajectories over multiple agents.
A common method is to reduce the full joint distribution of each combination of per-agent trajectories using global latent variables, i.e. learned global embeddings \citep{casas2020implicit, ngiam2022scene, girgis2021latent, zhou2023qcnext}.
\citet{jiang2023motiondiffuser} perform joint motion forecasting as denoising diffusion process and denoise sets of noisy trajectories conditioned on the scene context. 
\citet{seff2023motionlm} cast joint motion forecasting as language modeling using a vocabulary of discrete motion vectors and joint roll-outs for multiple agents.
These generative modeling approaches are limited to forecasts of two agents and exhibit high inference latency (e.g., 409\,ms for MotionDiffuser \citep{jiang2023motiondiffuser}).
In contrast, our method forecasts motion for up to 8 agents with significantly lower inference latency  ($\sim$60\,ms on A6000 Ada GPUs).

\textbf{Retrocausal forecasting models} incorporate a flow of information from later parts of predictions to earlier ones.
Consistent causal-retrocausal modeling \citep{zimmermann2012forecasting} extends RNNs with a retrocausal information flow, which is directed from later states to earlier states.
Consistent Koopman autoencoders \citep{azencot2020forecasting} retrocausally predict backward dynamics. 
In language modeling, bidirectional BERT models \citep{devlin2019bert, warner2024smarter} include attention mechanisms from later words in a sentence to earlier ones.
For motion forecasting, recent self-supervised pre-training methods \citep{cheng2023forecast, lan2024sept, wagner2024jointmotion} include retrocausal objectives, where earlier parts of trajectories are masked and reconstructed using later parts.
In contrast to our work, none of these methods explore using retrocausal mechanisms for issuing instructions.

\textbf{Instructable AI models}\footnote{Also referred to as instructable AI agents \citep{sun2024salmon, raad2024scaling}.} are designed to follow instructions effectively, with the goal of improving alignment \cite{sun2024salmon}, adaptability \cite{raad2024scaling}, or generalization \cite{guhur2023instruction, sarch2024helper, kimopenvla}.
\citet{ICLR2025_d8f17023} generate control vectors to interpret learned mechanisms and instruct models at inference.
Vision-and-language navigation (VLN) methods \citep{chen2021history, pashevich2021episodic} combine language and computer vision models to follow text-based navigation instructions. 
These methods are trained or fine-tuned to follow instructions using supervised learning.
Conversely, we show that regular training of motion forecasting leads to the ability to follow goal-based instructions and to adapt basic directional instructions to the scene context.

\textbf{Compressed trajectory representations}
are typically generated using lossy compression, which smoothes trajectories by removing high frequency components.
From a modeling perspective, compression lowers computational complexity.
Related methods compress trajectories using principal component analysis \cite{jiang2023motiondiffuser}, discrete cosine transforms \cite{mao2019learning}, eigenvalue decomposition \cite{bae2023eigentrajectory}, or Bézier curves \cite{hug2020introducing}.
Unlike \citet{mao2019learning}, we compress probabilistic representations of trajectories using discrete cosine transforms.  
Unlike \cite{jiang2023motiondiffuser, bae2023eigentrajectory}, our method is not data dependent, making it robust to dataset-specific noise.

\section{Method}
Our RetroMotion model forecasts multiple motion trajectories for each modeled agent.
A trajectory is a sequence of future positions ($x$- and $y$-coordinates) with positional uncertainties represented as probability densities. 
Using mixture distributions, our method decomposes motion forecasts in the following three ways. 

\subsection{Decomposing exponential power distributions}
We model the positional uncertainty of each trajectory point as density of a bivariate exponential power distribution
\begin{equation}
\begin{aligned}
\mathcal{D'}(x, y; \bm{\mu}, \bm{\sigma}, \beta) = \frac{\beta^2}{4 \sigma_x \sigma_y [\Gamma(1/\beta)]^2} \\
\cdot \exp\left(-\left|\frac{x-\mu_x}{\sigma_x}\right|^{\beta} 
- \left|\frac{y-\mu_y}{\sigma_y}\right|^{\beta} \right),  
\end{aligned}
\label{eq:exponential_power_density}
\end{equation}
where $\Gamma(\cdot)$ is the gamma function, $\bm{\mu} = (\mu_x, \mu_y)$ are mean or location parameters, $\bm{\sigma} = (\sigma_x, \sigma_y)$ are scale parameters, and $\beta$ is a shape parameter.

Platykurtic densities (i.e., with flat peaks) fail to concentrate probability mass near the predicted position, while heavy-tailed densities assign excessive probability to distant outcomes.
Both are undesirable in subsequent motion planning (see \cite{tacs2023decision, geisslinger2023ethical}).
Thus, we limit the shape parameter $\beta$ to the range of 1.0 to 2.0.
We approximate this as a mixture distribution of a bivariate normal ($\mathcal{N}$) and a bivariate Laplace ($\mathcal{L}$) distribution, with the following mixture density
\begin{equation}
    \begin{aligned}
    \mathcal{D}(x, y; w, \bm{\phi}) = w \cdot \mathcal{N}(x, y; \bm{\phi}) + (1 - w) \cdot \mathcal{L}(x, y; \bm{\phi}),
    \end{aligned}
\label{eq:approximated_exp_power_density}
\end{equation}
for learned weights $0 \le w \le 1$ and shared density parameters $\bm{\phi} = (\mu_x, \mu_y, \sigma_x, \sigma_y)$.
Following common practice \cite{nayakanti2023wayformer, zhou2023query, zhou2023qcnext}, we model the $x$- and $y$-coordinates as uncorrelated random variables.
In the following, we include $w$ in the tuple of density parameters
$\bm{\phi} = (w, \mu_x, \mu_y, \sigma_x, \sigma_y)$.

\subsection{Decomposing marginal motion forecasts}
\label{subsection:marginaldecoding}
We train a distinct decoder to perform marginal motion forecasting (i.e., per-agent).
Following common practice \cite{chai2020multipath, zhang2024real, nayakanti2023wayformer}, we predict the density of a mixture distribution at each future time step and fix mixture weights over time.
The corresponding mixture components describe positions of the same agent, but from different trajectories.
Following Bishop \cite{bishop1994mixture}, we express this as conditional probability
\begin{equation}
    \mathcal{P}_{t, 1:K}^\text{marginal}(\bm{y}\mid\bm{x}) = \sum_{k = 1}^K m_k(\bm{x}) \cdot \mathcal{D}\big(\bm{y}\mid\bm{\phi}_{t, k}(\bm{x})\big),
\end{equation}
where $\bm{x}$ is the input, $\bm{y}$ the target vector, $t \in \{1,...,T\}$ are future time steps, $K$ is the number of trajectories, $k$ indexes the corresponding mixture components, and $\bm{m}$ are mixture weights.
Unlike mixture weights, densities are variable across components and time steps.

\subsection{Decomposing joint motion forecasts}
\label{subsection:jointdecoding}
\begin{figure}[!t]
    \centering
    \includegraphics[width=0.9\linewidth]{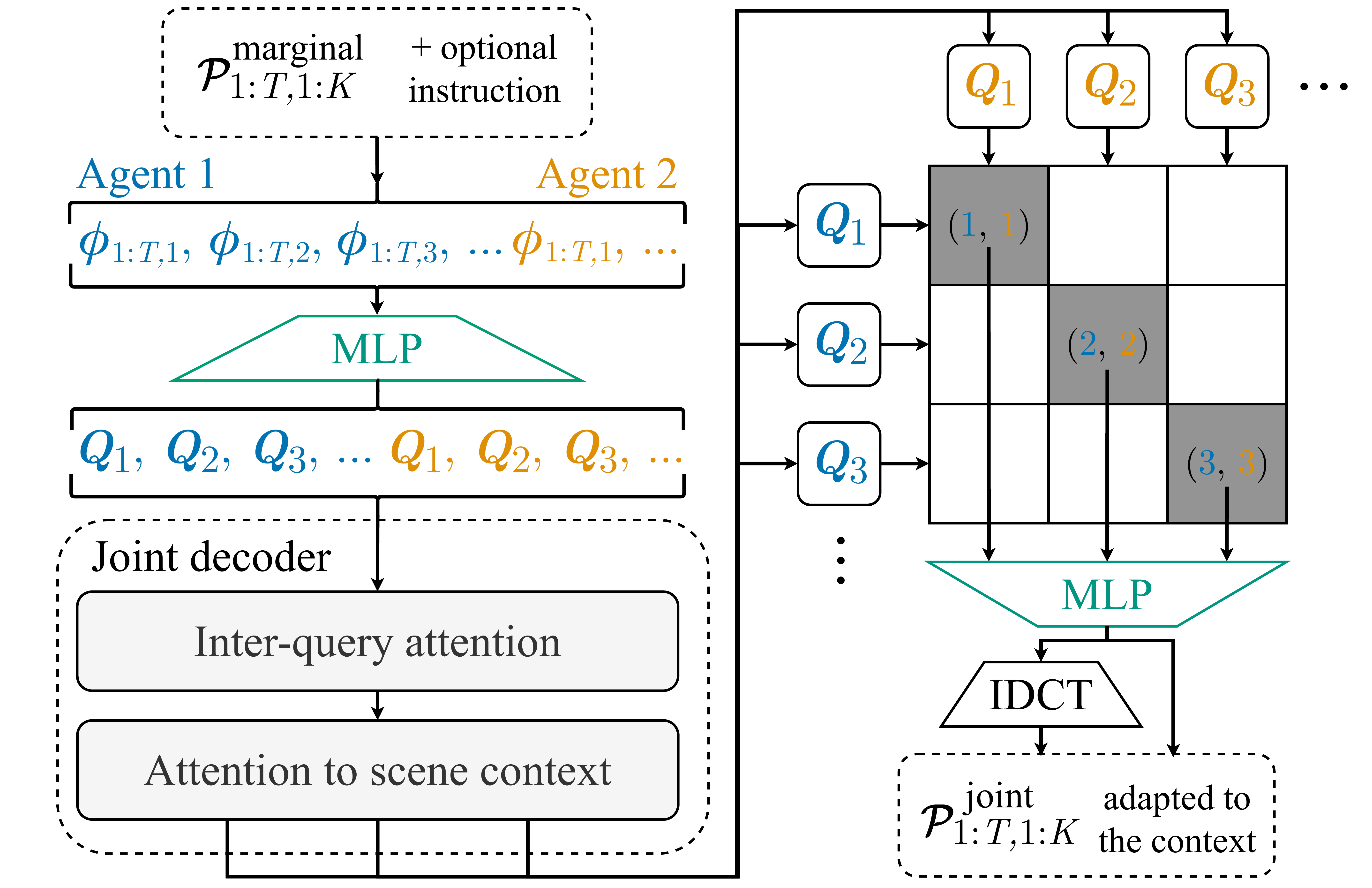}
    \caption{\textbf{From marginal to joint trajectory distributions.} 
     \textbf{Left part:} We use an MLP to generate query matrices $\bm{Q}$ from marginal trajectories and exchange information between queries and scene context.
     Scene context representations are learned by our scene encoder (see \Cref{subsection:sceneencoder}).
     \textbf{Right part:} Afterwards, we decode joint trajectories $\mathcal{P}^{\text{joint}}_{1:T}$ from pairs of queries at the same index for both agents ($(1, 1), (2, 2)$, ...). 
    This compresses information from all $K^2$ possible combinations into $K$ query pairs (full index matrix vs. its diagonal).
    Finally, we use the inverse discrete cosine transformation (IDCT) to decompress trajectories (\Cref{subsection:dct}).
    }
    \label{fig:per-agent-to-multi-agent}
\end{figure}

We generate joint trajectory distributions by re-encoding marginal distributions followed by pairwise modeling.
To effectively exchange information between agents, we transform all trajectories to the local reference frame of agent 1 and use the scene context embeddings of agent 1 (\Cref{subsection:sceneencoder}).
Afterwards, we exchange information via attention mechanisms and decode joint trajectory distributions using multi-agent mixture components (see \Cref{fig:per-agent-to-multi-agent}).
Per time step, each mixture component is a mixture density itself, describing one future position of each modeled agent
\begin{equation}
    \begin{aligned}
    \mathcal{P}_{t, 1:K}^\text{joint}(\bm{y}\mid\bm{x}) = \sum_{k = 1}^K c_k \sum_{a = 1}^A M_{k, a}(\bm{x})
    \cdot \mathcal{D}\big(\bm{y}\mid\bm{\phi}_{t, k, a}(\bm{x})\big),
    \end{aligned}
\end{equation}
where $A$ is the number of agents and $\bm{M}$ is a matrix of per-agent mixture weights.
We compute multi-agent mixture weights with 
\begin{equation}
    \bm{c} = \mathrm{softmax}\left(\frac{1}{\tau}\sum_{a=1}^A \bm{M}_{1:K,a}\right),
\label{eq:confidence}
\end{equation}
with a tunable temperature parameter $\tau$.

As shown in \Cref{fig:per-agent-to-multi-agent}, this approximates the joint distribution of all combinations of per-agent mixture components by focusing on the diagonal query pairs in matrix form.
Off-diagonal query pairs can update diagonal pairs through attention mechanisms. 
This compresses information from all $K^2$ possible combinations into $K$ query pairs.

For both marginal and joint motion forecasts, we follow related methods \cite{chai2020multipath, shi2022motion, zhou2023query} and use the out-most mixture weights ($m_k$ and $c_k$) as confidence scores.

\subsection{DCT compression of location parameters}
\label{subsection:dct}
Most motion forecasting methods regress trajectories at a frequency of 10\,Hz \citep{zhou2023query, zhang2024real, wagner2024redmotion}, allowing models to predict sudden changes between successive positions that are physically impossible yet close to the ground truth.
Such forecasts resemble noisy versions of smooth ground truth trajectories.
Therefore, we use a compressed probabilistic representation of trajectories without high frequency components.
Specifically, we incorporate the inverse discrete cosine transform (IDCT) into our model (see \Cref{fig:per-agent-to-multi-agent}) to internally represent density location parameters (i.e., means $\bm{\mu}$ in \Cref{eq:exponential_power_density} for all future time steps) as a sum of cosine functions. 
We hypothesize that this is a natural choice for transformer models given the use of sinusoidal positional encodings \citep{vaswani2017attention}.
To compress, we limit the frequencies in the IDCT to the lower end.
This method is data-independent, making it invariant to dataset or setup-specific noise (e.g., produced by errors of perception models).

\subsection{Scene encoder}
\label{subsection:sceneencoder}
We follow \citet{gao2020vectornet} and represent multimodal inputs (i.e., past trajectories, lane data, and traffic light states) as polylines.
We sample temporal features (past positions and traffic light states) with a frequency of 10\,Hz and static spatial features (lane markings and road borders) and a resolution of 0.5 meters.
We generate embeddings for each modality with 3-layer MLPs, add sinusoidal positional encodings, and process the embeddings with transformer encoder modules \cite{vaswani2017attention}.
Following \citet{nayakanti2023wayformer}, we initially process local agent-centric views within scenes (centered around each modeled agent) and compress them using cross-attention.
Afterwards, we change the batch dimension from the agent dimension to the scene dimension, and concatenate learned embeddings of Cartesian transformation matrices from agent-centric views into a global reference frame (as in \citep{jiang2023motiondiffuser}). 
Finally, we add global sinusoidal positional encodings and generate global scene context representations with further self-attention mechanisms.
Our two decoders (\Cref{subsection:marginaldecoding} and \Cref{subsection:jointdecoding}) decode probabilistic motion forecasts from this scene context.

\subsection{Loss function}
We train our model using maximum likelihood estimation with a multitask loss that covers the objectives described in \cref{subsection:marginaldecoding} and \cref{subsection:jointdecoding}.
Formally, we batchwise minimize the negative log-likelihood for forecasting the ground truth trajectories.
\begin{equation}
    \ell(\bm{x}, \bm{y}) = - \frac{1}{N} \sum^N_{n=1} \sum^T_{t=1} \ln\left(\mathcal{P}_{t, 1:K}^\text{joint}\right) + \lambda \ln\left(\mathcal{P}_{t, 1:K}^\text{marginal}\right),
\label{eq:lossfn}
\end{equation}
where $\bm{x}$ are inputs (see \Cref{subsection:sceneencoder}), $\bm{y}$ are targets (i.e., trajectories), $N$ is the number of samples in a batch, $\lambda$ is a tunable weighting factor.
Following \cite{shi2022motion, zhang2024real}, we optimize our objective by backpropagating only the loss for the trajectories that are closest to the ground truth trajectories.
We measure the distance to the ground truth using the $L_2$-distance (see \cite{zhang2024real}).
For marginal forecasts $\mathcal{P}_{1:T, 1:K}^\text{marginal}$, we select the best trajectory index $k$ for each agent individually.
For joint forecasts $\mathcal{P}_{1:T,1:K}^\text{joint}$, we select the best set of trajectories at the same index (see \Cref{fig:per-agent-to-multi-agent} right).
\section{Experiments}
We evaluate the motion forecasting performance of our method using the Waymo Open Motion \cite{ettinger2021large}, the Argoverse 2 Forecasting \cite{wilson2023argoverse}, and the V2X-Seq \cite{yu2023v2x} datasets.
Afterwards, we show that regular training of motion forecasting leads to the ability to follow goal-based instructions and to adapt basic directional instructions to the scene context.
Furthermore, we analyze learned density shapes and measure neural regression collapse \citep{andriopoulos2025prevalence} to interpret learned representations.

\subsection{Interactive motion forecasting}
\subsubsection{Dataset}
We use the Waymo Open Motion dataset to evaluate the joint motion forecasts of our model.
Specifically, we use the interactive test and validation splits to benchmark our model.
The dataset contains diverse traffic scenarios of urban and suburban driving and annotations of interacting road users.
All samples consist of 1 second of past agent states (i.e., position, velocity, acceleration, and bounding boxes) and map features such as road markings and traffic light states.
The prediction targets are future trajectories of up to 8 seconds.

\subsubsection{Model configurations}
\label{subsubsection:model_config}
We configure our marginal decoder to forecast marginal trajectory distributions of 8 agents and our joint decoder to forecast joint distributions for 2 agents per scenario.
For each agent, our model predicts 6 trajectories and the corresponding confidence scores.
Following Zhang et al. \cite{zhang2024real}, our scene encoder processes the 128 closest map polylines and up to 48 trajetories of surrounding agents per modeled agent.
We include an IDCT transform in our model that reconstructs 80 location parameters (for $x$- and $y$-coordinates) from 16 predicted DCT coefficients.
Overall, our model has 24\,M parameters.
As post-processing, we follow \citet{konev2022mpa} and reduce the confidence scores of redundant trajectories. 
We adapt this mechanism to joint trajectory sets by suppressing confidence scores only if a nearby trajectory belongs to a joint trajectory set with a higher accumulated confidence score (see $c$ in \Cref{eq:confidence}).
We determine the thresholds for each agent type using the training split (vehicles: 1.5\,m, pedestrians: 1.2\,m, cyclists: 1.2\,m).

Additionally, we build a sparse mixture of experts (SMoE) model \citep{jacobs1991adaptive, fedus2022review} of 3 variations of our model. 
All expert models are trained independently.
At inference, we use a rule-based router that selects one model based on the agent type.
Our RetroMotion (SMoE hybrid) configuration in \Cref{table:interactive_forecasting} is motivated by the observation that models with more anchors and static anchor initialization (like BeTopNet) perform better for cyclists.
Thus, we build a SMoE model that uses RetroMotion-based experts for vehicles and pedestrians, along with a reproduced BeTopNet for cyclists.

\subsubsection{Training details}
\label{subsubsection:training_details}
We sample 32 scenarios in a batch, with 8 focal (i.e., predicted) agents in each scenario.
Our model predicts marginal trajectory distributions for all 8 agents and joint distributions for two interactive agents.
We set $\lambda = 0.5$ (see \Cref{eq:lossfn}), use Adam with weight decay \cite{loshchilov2018decoupled} as the optimizer, and a step learning rate scheduler to halve the initial learning rate of $2^{-4}$ every 10 epochs.
We train RetroMotion for 50 epochs using data distributed parallel (DDP) training on 4 A6000 Ada GPUs.

\subsubsection{Motion forecasting metrics}
\label{subsubsection:forecasting_metrics}
The official challenge metrics are the mean average precision (mAP), the average displacement error (minADE), and the final displacement error (minFDE), the miss rate (MR), and the overlap rate (OR).
All metrics are computed using the joint trajectory set closest to the ground truth trajectories.
Furthermore, the metrics are averaged over the three prediction horizons of 3\,s, 5\,s, and 8\,s, and the 3 agent types (i.e., vehicles, pedestrians, and cyclists).
For further details, refer to Ettinger et al. \cite{ettinger2021large}.

\subsubsection{Results}
\Cref{table:interactive_forecasting} presents motion forecasting metrics for interactive forecasting on the Waymo Open Motion dataset.
The QCNeXt model \cite{zhou2023qcnext} performs strongly on the exclusively distance-based metrics like minFDE, indicating that our model predicts higher confidence scores for trajectories close to the ground truth.
Averaged over all agent types, MTR++, QCNeXt, BeTopNet, and our RetroMotion model are within one mAP point.
Our SMoE configurations reach the highest mAP scores, with slightly worse minFDE scores compared to our base model.

\begin{figure*}[]
    \centering
    \begin{subfigure}[t]{0.246\textwidth}
        \centering
        \includegraphics[width=\linewidth]{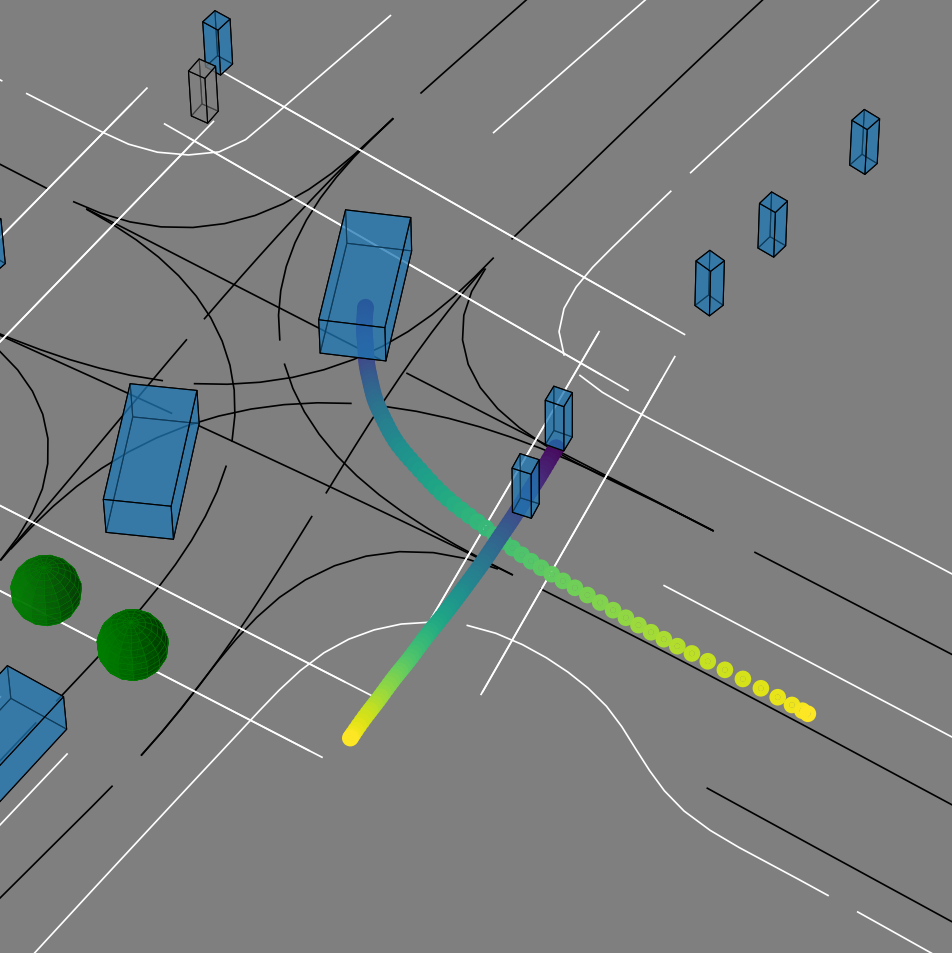}
        \caption{A car yielding for pedestrians}
    \end{subfigure}
    \begin{subfigure}[t]{0.246\textwidth}
        \centering
        \includegraphics[width=\linewidth]{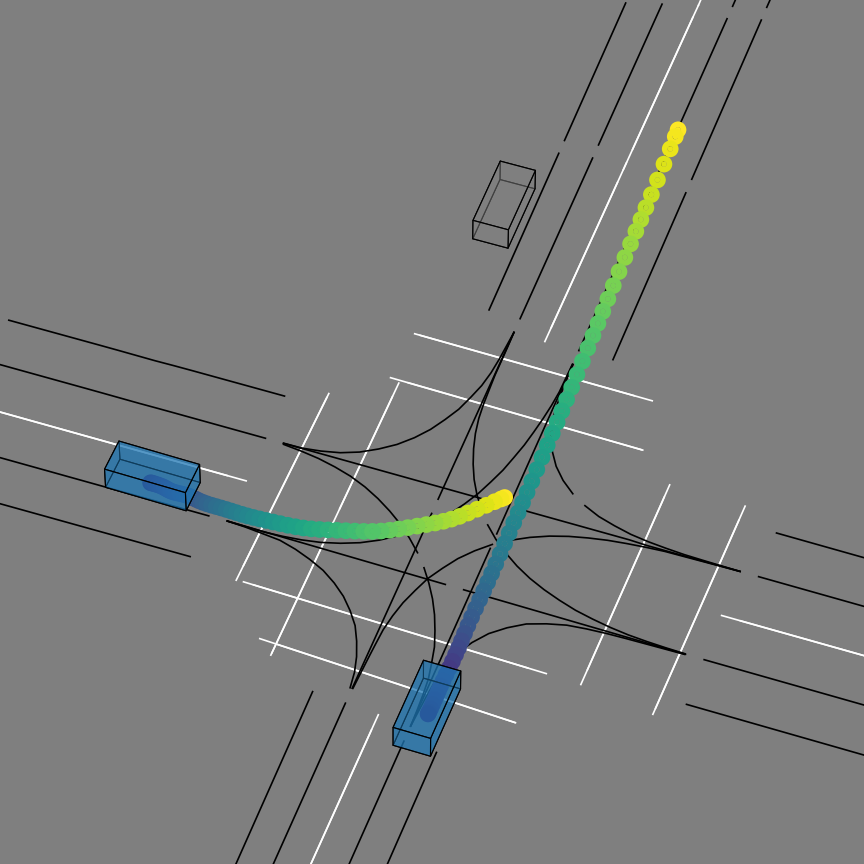}
        \caption{Right of way is given}
    \end{subfigure}
    \begin{subfigure}[t]{0.246\textwidth}
        \centering
        \includegraphics[width=\linewidth]{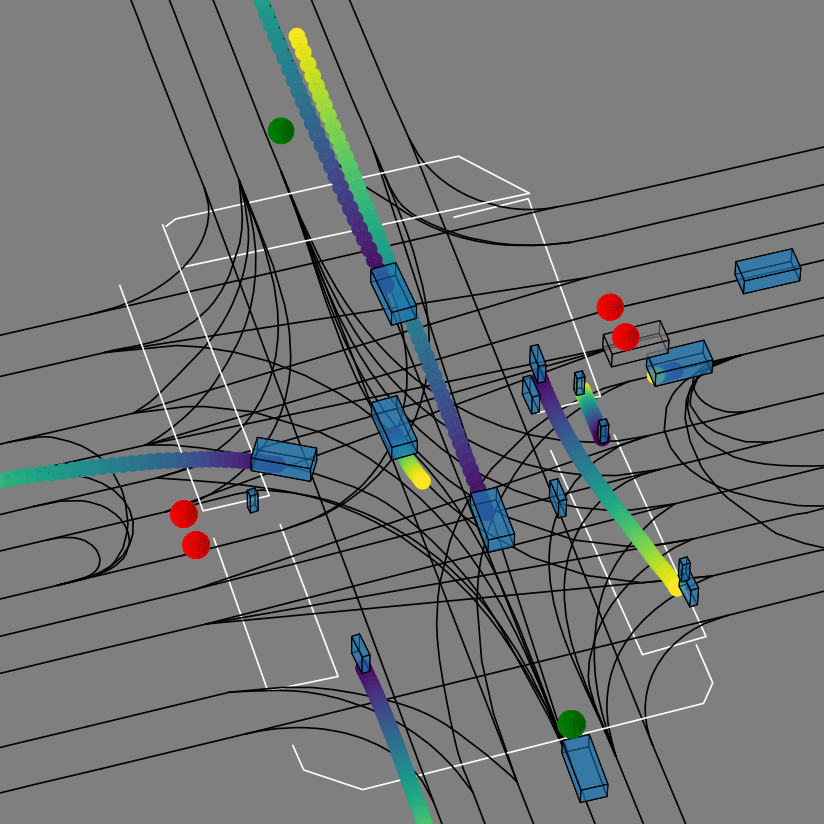}
        \caption{Complex scenario}
    \end{subfigure}
    \begin{subfigure}[t]{0.246\textwidth}
        \centering
        \includegraphics[width=\linewidth]{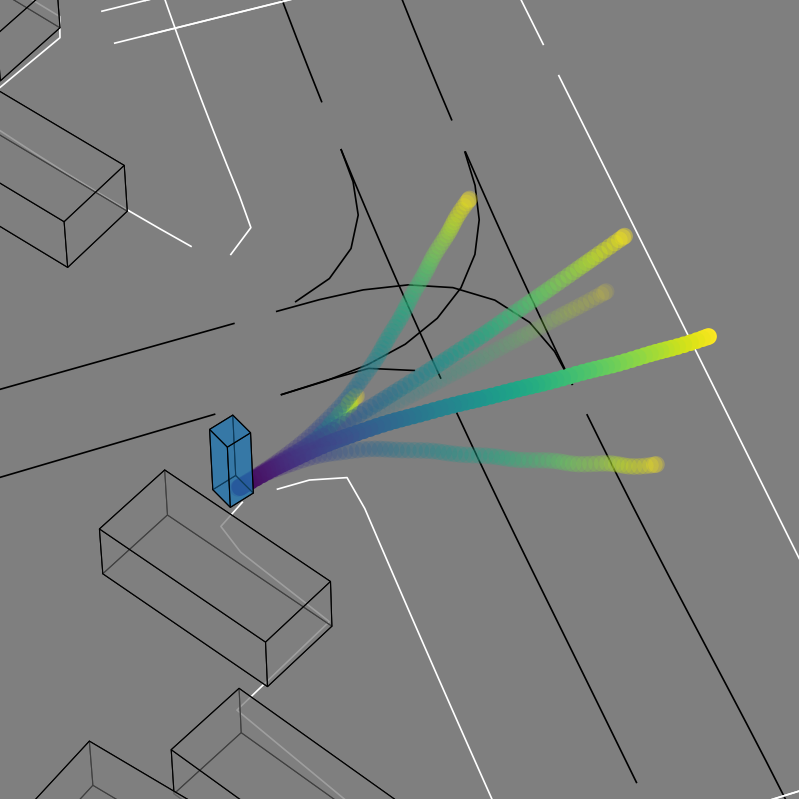}
        \caption{6 marginal modes}
    \end{subfigure}

    \caption{\textbf{Joint and marginal motion forecasts of our model.} Dynamic agents are shown in blue, static agents in grey (determined at $t=0\,\text{s}$). Lanes are black lines, road markings are white lines, and traffic light states are shown as colored spheres. (a) and (b): Top1 mode of joint motion forecasts on the Waymo Open Motion and Argoverse 2 datasets. (c) Marginal forecasts on the V2X-Seq dataset. \mbox{(d) 6 modes} of a marginal forecast for a pedestrian, the opacity is based on the confidence score fore each mode. }
    \label{figure:qualitative}
\end{figure*}

Furthermore, we follow \citet{ngiam2022scene} and evaluate the marginal predictions of our marginal decoder as joint predictions on the validation split (see the  marginal as joint configuration in \Cref{table:interactive_forecasting}).
All forecasting metrics are significantly worse than for the joint predictions.
This highlights the ability of our model to perform joint modeling by re-encoding and adapting the marginal predictions to the predictions of surrounding agents. 
\Cref{figure:qualitative} shows qualitative results of our method. 
We show joint and marginal forecasts to highlight that our method is suitable for more complex scenarios with many agents.

\begin{table}[!h]
\setlength{\tabcolsep}{10pt}
\centering
\resizebox{0.48\textwidth}{!}{
    \begin{tabular}{lccccc}
    \toprule
    Method (config) & mAP\,$\uparrow$  & minFDE\,$\downarrow$ & OR\,$\downarrow$ \\ \midrule
         \textit{Test split} \\
         Scene Transformer (joint) \citep{ngiam2022scene} & 0.1192 & 2.1892 & 0.2067 \\ 
         GameFormer (joint) \citep{huang2023gameformer} & 0.1376 & \underline{1.9373} & 0.2112 \\
         SceneMotion (joint) \citep{wagner2024scenemotion} & 0.1789 & 2.3141 & 0.2163 \\
         JointMotion (HPTR) \citep{wagner2024jointmotion} & 0.1869 & 2.0507 & 0.2037 \\
         MotionDiffuser \citep{jiang2023motiondiffuser} & 0.1952 & 1.9482 & 0.2004 \\
         JFP \citep{luo2023jfp} & 0.2050 & 1.9905 & 0.1835 \\
         MotionLM \citep{seff2023motionlm} & 0.2178 & 2.0067 & 0.1823 \\
         MTR++ \citep{shi2024mtr++} & 0.2326 & 1.9509 & \textbf{0.1665} \\
         QCNeXt \citep{zhou2023qcnext} & 0.2352 & \textbf{1.6772} & 0.1946 \\
         BeTopNet \citep{liu2024betop} & 0.2412 & 2.2744 & \underline{0.1695} \\
         RetroMotion [ours] & 0.2397 & 1.9591 & 0.2020 \\
         RetroMotion (SMoE) [ours] & \underline{0.2422} & 2.0245 & 0.2007 \\
         RetroMotion (SMoE hybrid) [ours] & \textbf{0.2519} & 2.0890 & 0.1927 \\
         \midrule
         \textit{Validation split} \\
         MotionLM (single replica) \citep{seff2023motionlm} & 0.1687 & 2.3886 & - \\
         MTR++ \citep{shi2024mtr++} & \underline{0.2398} & \underline{1.9712} & - \\
         RetroMotion [ours] & \textbf{0.2441} & \textbf{1.8728} & 0.2019 \\
         RetroMotion (marginal as joint) & 0.1797 & 2.0118 & 0.2200 \\
    \bottomrule
    \end{tabular}
}
\caption{\textbf{RetroMotion performs well on interactive scenarios.} All methods are evaluated on the interactive splits of the Waymo Open Motion \mbox{Dataset \citep{ettinger2021large}}. The main metric is the mAP. 
As \textit{marginal as joint} configuration (see \cite{ngiam2022scene}), we evaluate the initial marginal predictions as joint predictions.
The results are significantly worse in this configuration, which highlights the improvements corresponding to our second joint decoder.
Best scores are \textbf{bold}, second best are \underline{underlined}.}
\label{table:interactive_forecasting}
\end{table}

\subsection{Cross-dataset generalization}
We follow UniTraj \cite{feng2024unitraj} to measure the cross-dataset generalization capabilities of our method.
Thus, we configure our model to forecasts 6s-long trajectories for evaluation on Argoverse 2 Forecasting \cite{wilson2023argoverse} and 5s-long trajectories for evaluation on V2X-Seq \cite{yu2023v2x}.
Besides that, we train our model on the Waymo dataset with the same settings as in \Cref{subsubsection:model_config} and \Cref{subsubsection:training_details}.

\subsubsection{Cross-dataset evaluation metrics}
\label{subsubsection:cross_dataset_metrics}
UniTraj focuses on marginal forecasting, thus we evaluate the marginal predictions of our model.
UniTraj uses the same distance-based metrics described in \Cref{subsubsection:forecasting_metrics}, but evaluated on an agent-by-agent (i.e., marginal) basis.
Additionally, we report the $\text{Brier-minFDE} = \text{minFDE} + (1 - p)^2$, where $p$ is the predicted confidence score per trajectory.
Following UniTraj, we report the metrics for vehicles.
For the V2X-Seq dataset, we average the metrics across the 3 splits (i.e., vehicle-view, infrastructure-view, and cooperative-view).

\subsubsection{Results}
\Cref{table:cross_dataset} shows the results of this experiment.
On the Argoverse 2 Forecasting dataset, our method outperforms MTR \cite{shi2022motion} and Wayformer \cite{nayakanti2023wayformer} and competes with AutoBot \cite{girgis2021latent} (lower miss rate, but higher minFDE scores).
On the V2X-Seq dataset, RetroMotion outperforms Wayformer \cite{nayakanti2023wayformer} and RedMotion \cite{wagner2024redmotion}.
Overall, these results show that RetroMotion generalizes well across dataset.

\begin{table}[!h]
\setlength{\tabcolsep}{10pt}
\centering
\resizebox{0.47\textwidth}{!}{
    \begin{tabular}{llccc}
    \toprule
    Dataset &  Method &  Brier-minFDE\,$\downarrow$ & minFDE\,$\downarrow$ & MR\,$\downarrow$  \\ \midrule
    \multirow{4}{2cm}{Argoverse 2 \\ Forecasting} & MTR \citep{shi2022motion} * & 3.63 & 3.14 & 0.44 \\
    & Wayformer \citep{nayakanti2023wayformer} * & 3.60 & 3.14 & 0.45 \\
    & AutoBot \citep{girgis2021latent} * & \textbf{3.23} & \textbf{2.41} & \underline{0.40} \\
    & RetroMotion [ours] & \underline{3.51} & \underline{2.84} & \textbf{0.31} \\
    \midrule
    \multirow{3}{*}{V2X-Seq} & Wayformer \cite{nayakanti2023wayformer} & 2.36 & 1.95 & 0.34 \\
    & RedMotion \cite{wagner2024redmotion} & \underline{2.32} & \underline{1.90} & \underline{0.30} \\
    & RetroMotion [ours] & \textbf{2.04} & \textbf{1.56} & \textbf{0.21} \\
    \bottomrule
    \end{tabular}
}
\caption{\textbf{RetroMotion generalizes from Waymo Open Motion to Argoverse 2 Forecasting and V2X-Seq.} All methods are trained on Waymo Open Motion. Best scores are \textbf{bold}, second best are \underline{underlined}. *Results reported in \cite{feng2024unitraj}.}
\label{table:cross_dataset}
\end{table}

\subsection{Issuing instructions by modifying trajectories}
In the following, we test our model's ability to follow goal-based instructions and to adapt basic directional instructions to the scene context.
Specifically, we issue instructions by modifying predicted marginal trajectories prior to re-encoding and joint modeling (see \Cref{fig:per-agent-to-multi-agent}).

\subsubsection{Goal-based instructions}
In this experiment, we use the last second of the ground truth trajectories as goal-based instructions.
Specifically, we replace the last second of predicted marginal trajectories with the last second of ground truth trajectories and evaluate the changes in joint trajectories ($\mathcal{P}^{\text{joint}}_{1:T}$ in \Cref{fig:per-agent-to-multi-agent}).
For evaluation, we use the validation split of the Argoverse 2 Forecasting dataset and the metrics described in \Cref{subsubsection:cross_dataset_metrics}.
This evaluation is similar to the \textit{goal-conditioned} configuration of \citet{ngiam2022scene}, but with trajectory modifications as goal-based instructions instead of goal positions as input.
At inference, we modify either the marginal trajectory with the highest confidence score or with the lowest minFDE w.r.t. the desired goal positions, or all 6 trajectories per agent.

\textbf{Results:}
All instruction configurations are evaluated on the validation split of the Argoverse 2 Forecasting dataset.
\Cref{table:goal_based_instructions} presents the results.
All instruction configurations improve the distance-based metrics, while modifying all trajectories leads to the most significant improvement (12\% lower final distance error).
This shows that modifications to marginal trajectories affect subsequently decoded joint trajectories and can be used to issue goal-based instructions.

\begin{table}[!h]
\setlength{\tabcolsep}{10pt}
\centering
\resizebox{0.47\textwidth}{!}{
    \begin{tabular}{llcc}
    \toprule
    Instruction config. & minFDE\,$\downarrow$  & minADE\,$\downarrow$ & MR\,$\downarrow$  \\ \midrule
    None & 2.45 & 1.25 & 0.31 \\
    Highest confidence & 2.41 \, -1.6\% & 1.24 & 0.31 \\
    Lowest minFDE & \underline{2.28} \, -6.9\% & \underline{1.19} & \underline{0.29} \\
    All trajectories & \textbf{2.16} \, -11.8\% & \textbf{1.15} & \textbf{0.27} \\
    \bottomrule
    \end{tabular}
}
\caption{\textbf{RetroMotion follows goal-based instructions.} As instructions, we modify marginal trajectories and evaluate the changes in joint trajectories. 
We modify either the trajectory with the highest confidence score or with the lowest minFDE w.r.t. the desired goal positions, or all 6 trajectories.
All configurations are evaluated on the validation split of the Argoverse 2 Forecasting \mbox{dataset \cite{wilson2023argoverse}}. Best scores are \textbf{bold}, second best are \underline{underlined}.}
\label{table:goal_based_instructions}
\end{table}

\subsubsection{Basic directional instructions}
\label{subsubsection:basic_directional_instructions}
Building on the insight that our model can follow goal-based instructions, we further evaluate basic directional instructions.
We define basic directional instructions for turning left and right.

\begin{figure}
    \centering
    \includegraphics[width=0.95\linewidth]{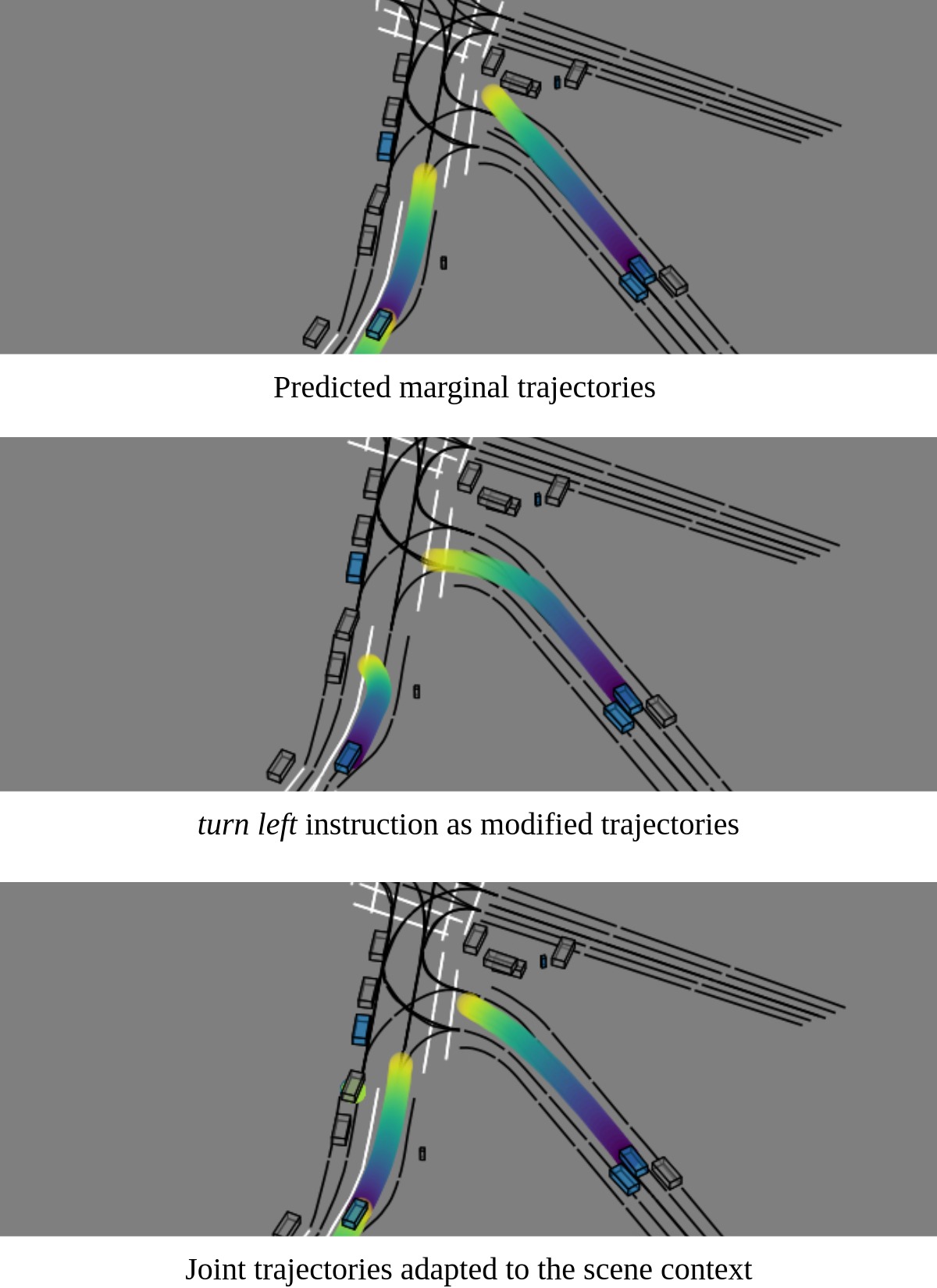}
    \caption{
    \textbf{Adapting a basic turn left instruction to the scene context.}
    The upper plot shows the default marginal trajectory forecast of our model.
    The middle plot shows our basic \textit{turn left} instructions, which violate traffic rules by turning into the oncoming lanes.
    The lower plot shows that our model responds to this instruction by adapting the trajectory of the right vehicle to its lane (shown as black line) and   reversing the instruction for the left vehicle, since turning left is not possible.
    }
    \label{fig:directional}
\end{figure}

We describe both instructions with trajectories based on quarter circles.
Specifically, we scale the radius $r$ of the circle to maintain an agent's current speed. 
Then we use the upper right quadrant, shifted to the origin, as a \textit{turn left} instruction, with
\begin{equation}
\begin{aligned}
    x(t) = r \cdot \cos(t) - r, \\
y(t) = r \cdot \sin(t).
\end{aligned}
\end{equation}
Similarly, we use the upper left quadrant shifted to the origin as the \textit{turn right} instruction.  
At inference, we issue the instructions by replacing the last 4 seconds in all marginal trajectories with the corresponding quarter circle \mbox{(see \Cref{fig:directional})}.

Note these instructions are intentionally not adapted to the scene context (map and other agents).
In this experiment, we evaluate the ability of RetroMotion to adapt our directional instructions to the given context.
Specifically, we measure the overlap rate with other agents \cite{ettinger2021large} of the modified trajectories used as instructions versus the subsequently decoded joint trajectories.
Furthermore, we compute average on-road probability (ORP) scores, which describe whether trajectories are staying on the road versus going off-road\footnote{We frame on-road prediction as a binary classification task: a trajectory is labeled off-road if any of its waypoints fall outside the driveable area, and on-road otherwise.}. 

\textbf{Results:}
\Cref{table:directional_instructions} presents the results of this experiment.
Overall, higher ORP scores and lower OR scores are obtained for the \textit{turn right} than for the \textit{turn left} instructions. 
We hypothesize that this is due to the fact that the dataset mainly contains right-hand traffic scenarios, where right turns are commonly allowed and more frequent. 
Notably, the adjusted joint trajectories have significantly lower OR scores (22\% and 14\% lower) and much higher ORP scores (33\% and 20\% higher).
This highlights the ability of our model to adapt basic directional instructions to the given scene context.

\begin{table}[!h]
\setlength{\tabcolsep}{10pt}
\centering
\resizebox{0.47\textwidth}{!}{
    \begin{tabular}{llll}
    \toprule
    Instruction & Eval. trajectory & OR\,$\downarrow$ & ORP\,$\uparrow$  \\ \midrule
    \multirow{2}{*}{\textit{turn left}} &  basic instruction & 0.23 & 0.64 \\
    & adapted joint traj. & 0.18 \, -22\% & 0.85 \, +33\% \\
    \midrule
    \multirow{2}{*}{\textit{turn right}} &  basic instruction & 0.21 & 0.76 \\
    & adapted joint traj. & 0.18 \, -14\% & 0.91 \, +20\% \\
    \bottomrule
    \end{tabular}
}
\caption{\textbf{RetroMotion adapts directional instructions to the scene context.} As instructions, we modify marginal trajectories and evaluate the changes in joint trajectories.
We report overlap rates (OR) with other agents and on-road probability (ORP) scores.
All configurations are evaluated on the validation split of the Argoverse 2 Forecasting dataset.}
\label{table:directional_instructions}
\end{table}

\subsection{Analyzing learned trajectory representations}
\label{subsection:analyzing_reps}
In order to analyze the learned representations and perform an ablation study on our design choices, we train different configurations of our model.
Specifically, we ablate modeling positional uncertainty with different probability distributions, including normal, Laplace, and exponential power distributions. We also train our model with and without compressing the location parameters of the distributions.

We train each model configuration for 14 epochs on the Waymo Open Motion dataset and keep the remaining configurations as in \Cref{subsubsection:model_config} and \Cref{subsubsection:training_details}. 
\Cref{table:ablation} presents the results of this experiment.
Using Laplace distributions to model positional uncertainty improves motion prediction metrics over using normal distributions.
Compressing the location parameters of densities further improves the results significantly. 
Overall, using exponential power distributions with DCT compression leads to the best results across all metrics.

\begin{table}[!h]
\setlength{\tabcolsep}{10pt}
\centering
\resizebox{0.47\textwidth}{!}{
    \begin{tabular}{llccc}
    \toprule
    Distribution & DCT &  mAP\,$\uparrow$ & minFDE\,$\downarrow$  & minADE\,$\downarrow$   \\ \midrule
    Normal & False & 0.172 & 2.177 & 0.954 \\
    Laplace & False & 0.176 & 2.149 & 0.940 \\
    Laplace & True & \underline{0.194} & \underline{2.102} & \underline{0.917} \\
    Exponential power & True & \textbf{0.195} & \textbf{2.060} & \textbf{0.910} \\
    \bottomrule
    \end{tabular}
}
\caption{\textbf{Modeling positional uncertainty with exponential power distributions outperforms using normal or Laplace distributions.}
All configurations are evaluated on the interactive validation split of the Waymo Open Motion dataset. Best scores are \textbf{bold}, second best are \underline{underlined}.}
\label{table:ablation}
\end{table}

\subsubsection{Weights in exponential power distributions}
During training, we track the value of the mixture weights of normal components in exponential power distributions (see \Cref{figure:normal_weight}).
\begin{figure}[h!]

\pgfplotstableread[col sep=comma,]{data/pred_expp_gate.csv}\pred
\pgfplotstableread[col sep=comma,]{data/pred_0_expp_gate.csv}\predzero

\centering
{\resizebox{0.95\columnwidth}{!} {
    \begin{tikzpicture}

    \begin{axis}[
      no markers, domain=0:10, xmin=0, ymax=0.15,
      axis lines*=left, xlabel={Epoch}, ylabel=$w$,
      xtick scale label code/.code={},
      every axis x label/.style={at=(current axis.right of origin),anchor=west},
      height=5cm, width=12cm,
      enlargelimits=false, clip=false, axis on top,
      grid=major, grid style={dashed,gray!30},
      thick,
      legend pos=north west
      ]
      \addplot[orange] table [x=epoch, y=exppowergate]{\pred};
      \addplot[blue] table [x=epoch, y=exppowergate]{\predzero};
     
     \legend{joint, marginal}
    
    \end{axis}
    \end{tikzpicture}
}}

\caption{\textbf{Mixture weight of normal components in exponential power distributions (see \Cref{eq:approximated_exp_power_density}).}
The weight $w$ progressively increases, reaching higher values for joint trajectory distributions than for marginal ones. 
However, $w$ remains below 0.15, indicating that the learned distributions are Laplace-like.
}
\label{figure:normal_weight}
\end{figure}

Initially, the weight is close to 0, because the negative log-likelihood (NLL) of a normal distribution is closely related to the mean squared error, while the NLL of a Laplace distribution is related to the mean absolute error.
Therefore, the NLL of a normal distribution is much higher for outliers (i.e., unreasonable predictions during the earlier training epochs), which is compensated by a low $w$ value.
During training, $w$ progressively increases, while reaching higher values for predicted joint trajectory distributions than for marginal distributions. 
However, on average, the learned exponential power distributions tend to be Laplace-like with normal components with relatively low weights of 0.1 (joint) and 0.04 (marginal).

\subsubsection{Neural regression collapse}
We measure the neural regression collapse metric NRC1 \cite{andriopoulos2025prevalence} for feature vectors of marginal and joint trajectory distributions.
These feature vectors correspond to the last hidden states, which are generated by the MLP heads in our model.
We compute the metric assuming 272 or 32 principal components, as there are 272 density parameters per trajectory, including location parameters compressed as 32 DCT coefficients.
The upper plot in \Cref{figure:nrc1} shows that the 512-dimensional feature vectors collapse to a lower-dimensional subspace spanned by fewer than 272 principal components.
The lower plot shows no such collapse. 
Therefore, the true dimensionality of the feature vectors lies between 32 and 272.
These results suggest that, in addition to DCT coefficients for location parameters, density parameters such as shape and scale are also important when regressing trajectories (see \Cref{eq:exponential_power_density}).

\begin{figure}[h!]

\pgfplotstableread[col sep=comma,]{data/pred_nrc1.csv}\pred
\pgfplotstableread[col sep=comma,]{data/pred_0_nrc1.csv}\predzero

\centering
{\resizebox{0.95\columnwidth}{!} {
    \begin{tikzpicture}

    \begin{axis}[
      no markers, domain=0:10, xmin=0, ymax=0.00025,
      axis lines*=left, xlabel={Epoch}, ylabel=NRC1 dim 272,
      xtick scale label code/.code={},
      every axis x label/.style={at=(current axis.right of origin),anchor=west},
      height=5cm, width=12cm,
      enlargelimits=false, clip=false, axis on top,
      grid=major, grid style={dashed,gray!30},
      thick,
      ]
      \addplot[orange] table [x=epoch, y=nrc1]{\pred};
      \addplot[blue] table [x=epoch, y=nrc1]{\predzero};
     
     \legend{joint, marginal}
    
    \end{axis}
    \end{tikzpicture}
}}

\pgfplotstableread[col sep=comma,]{data/pred_nrc1_32.csv}\predthree
\pgfplotstableread[col sep=comma,]{data/pred_0_nrc1_32.csv}\predzerothree

{\resizebox{0.95\columnwidth}{!} {
    \begin{tikzpicture}

    \begin{axis}[
      no markers, domain=0:10, xmin=0, ymax=0.03,
      axis lines*=left, xlabel={Epoch}, ylabel=NRC1 dim 32,
      xtick scale label code/.code={},
      every axis x label/.style={at=(current axis.right of origin),anchor=west},
      height=5cm, width=12cm,
      enlargelimits=false, clip=false, axis on top,
      grid=major, grid style={dashed,gray!30},
      thick,
      legend pos=south east
      ]
      \addplot[orange] table [x=epoch, y=nrc1]{\predthree};
      \addplot[blue] table [x=epoch, y=nrc1]{\predzerothree};
     
     \legend{joint, marginal}
    
    \end{axis}
    \end{tikzpicture}
}}

\caption{\textbf{Neural regression collapse for motion forecasting.} 
We measure the NRC1 metric for feature vectors of marginal and joint trajectory distributions.
There is an immediate collapse in the upper plot, but none in the lower plot. Therefore, the true dimensionality lies between 32 and 272.
This suggests that other density parameters besides the 32 location parameters are important.
}
\label{figure:nrc1}
\end{figure}
\section{Conclusion}
In this work, we  decompose the multi-agent motion forecasting task into modeling marginal trajectory distributions for all modeled agents and joint distributions for interacting agents.
This approach captures the varying complexity inherent in different traffic scenarios while maintaining computational efficiency.
Our transformer-based forecasting model incorporates a retrocausal information flow and models positional uncertainty through compressed exponential power distributions.
This enables more accurate predictions across diverse scenarios.


Furthermore, our method's ability to respond to goal-based and basic directional instructions reveals an emergent capability that was not explicitly trained for. 
This suggests that standard motion forecasting training naturally develops representations that can interpret and adapt to goal-based and directional instructions within the appropriate context. 
This capability can improve human-AI interaction in self-driving systems, potentially allowing operators to guide vehicle behavior.

Future work can further explore the instruction-following capabilities, investigating the limits of generalization to more complex scenarios, and integrating this approach with motion planning modules. 
Overall, our decomposition-based method represents a significant step toward more accurate, interpretable, and interactive motion forecasting systems for self-driving vehicles.
\section*{Acknowledgments}
This paper includes results of chapter 5 of Royden Wagner's doctoral dissertation, 'Interpretable Representation Learning for Motion Forecasting' \cite{Wagner2026_1000190386}. \newline
The research leading to these results is partially funded by the German Federal Ministry for Economic Affairs and Energy Action within the project ``NXT GEN AI METHODS''.
The authors gratefully acknowledge the computing time provided on the high-performance computer HoreKa by the National High-Performance Computing Center at KIT (NHR@KIT). This center is jointly supported by the Federal Ministry of Education and Research and the Ministry of Science, Research and the Arts of Baden-Württemberg, as part of the National High-Performance Computing (NHR) joint funding program. HoreKa is partly funded by the German Research Foundation (DFG).

{
    \small
    \bibliographystyle{ieeenat_fullname}
    \bibliography{main}
}

\end{document}